\documentclass[12pt]{article}

\usepackage{graphicx}
\usepackage{times}
\usepackage{comment}
\usepackage{amsmath}
\usepackage{lineno}

\usepackage{tabularx}
\usepackage{booktabs}
\usepackage{ragged2e}
\usepackage{microtype}
\usepackage{subfig}
\usepackage{hyperref}
\usepackage{cleveref}

\usepackage{amssymb}   % for \checkmark
\usepackage{textcomp}  % for \texttimes
\usepackage{array}
\usepackage{makecell}  % for line breaks and horizontal line in cell
\usepackage{diagbox}   % for diagonal top-left cell
\usepackage{xcolor}     % for coloring
\usepackage{lipsum}
\usepackage{caption}

\newenvironment{sciabstract}{%
\begin{quote} \bf}
{\end{quote}}

\title{Flocking phase transition and threat responses in bio-inspired autonomous drone swarms}

\author
{Matthieu Verdoucq,$^{1, 2, \dagger}$ Dari Trendafilov,$^{2, 1, \dagger}$ \protect\\ Clément Sire,$^{3}$ Ramón Escobedo$^{2,3}$, \protect\\ Guy Theraulaz$^{2}$, Gautier Hattenberger$^{1, \ast}$\\
\\
\normalsize{$^{1}$École Nationale de l’Aviation Civile, Université de Toulouse, France}\\
\normalsize{$^{2}$Centre de Recherches sur la Cognition Animale, Centre de Biologie Intégrative (CBI),}\\
\normalsize{Centre National de la Recherche Scientifique (CNRS) \& Université de Toulouse -- Paul Sabatier, France}\\
\normalsize{$^{3}$Laboratoire de Physique Théorique, CNRS \& Université de Toulouse -- Paul Sabatier, France}\\
\\
\normalsize{$^\dagger$ These authors contributed equally to this work.}\\
\normalsize{$^\ast$Corresponding author: gautier.hattenberger@enac.fr}
}

\date{}

\begin{document} 

\maketitle 

\begin{sciabstract}
{\large Abstract} \vspace{0.2cm}\\
Collective motion inspired by animal groups offers powerful design principles for autonomous aerial swarms. We present a bio-inspired 3D flocking algorithm in which each drone interacts only with a minimal set of influential neighbors, relying solely on local alignment and attraction cues. By systematically tuning these two interaction gains, we map a phase diagram revealing sharp transitions between swarming and schooling, as well as a critical region where susceptibility, polarization fluctuations, and reorganization capacity peak. Outdoor experiments with a swarm of ten drones, combined with simulations using a calibrated flight-dynamics model, show that operating near this transition enhances responsiveness to external disturbances. When confronted with an intruder, the swarm performs rapid collective turns, transient expansions, and reliably recovers high alignment within seconds. These results demonstrate that minimal local-interaction rules are sufficient to generate multiple collective phases and that simple gain modulation offers an efficient mechanism to adjust stability, flexibility, and resilience in drone swarms.
\end{sciabstract}

\newpage

\section*{Introduction}

Collective motion in natural systems offers a powerful source of inspiration for designing large teams of autonomous aerial robots capable of acting coherently in dynamic and unpredictable environments \cite{tahir2019}. Animal groups such as fish schools and bird flocks achieve remarkable feats of coordination using only local sensing, decentralized decision-making, and simple behavioral rules \cite{Radakov1973,ParrishHamner1997,Lopez2012,HerbertRead2016}. They reconfigure rapidly, maintain cohesion without collisions, and respond collectively to threats even when environmental information is sparse or noisy \cite{Treherne1981,Rieucau2014,Procaccini2011,Doran2022}. These biological systems have increasingly inspired distributed control laws for Unmanned Aerial Vehicle (UAV) swarms, with the ambition of replicating their scalability, robustness, and adaptability in real-world missions \cite{Hildmann2019,Zhou2020,Zhang2022,deCroon2023IntelligentSwarms}. However, despite significant technological progress, a persistent gap remains between the elegant predictions of theoretical flocking models and the performance of physical robot swarms operating under sensing noise, communication delays, and aerodynamic disturbances. Reducing this gap is essential for deploying UAV collectives in field scenarios ranging from monitoring rapidly evolving environmental hazards to autonomous surveillance and search and rescue, or defense against adversarial intruders.

Recent surveys highlight the multifaceted nature of this challenge. Scaling from single-robot autonomy to multi-robot collectives requires the tight integration of sensing, control, communication, and onboard computation, especially for aerial platforms that are strongly constrained by size–weight–power limitations and must manage fully three-dimensional flight dynamics~\cite{chung2018}. At the level of individual vehicles, the achievable collective behavior of micro-air-vehicle swarms is tightly limited by the reliability and precision of onboard sensing and state estimation, which motivates flocking algorithms that can operate with minimal and noisy information~\cite{coppola2020}. From a systems perspective, turning multiple UAVs into a functioning multi-drone system depends critically on ensuring sufficient connectivity, robust communication, and appropriate coordination schemes that can cope with changing mission goals and environmental disturbances~\cite{rinner2021}.

Large-scale field experiments have begun to show that bio-inspired decentralized control can be realized outside laboratory conditions. Outdoor experiments with multi-copter swarms have demonstrated that local flocking rules combined with minimal GPS-based relative information can sustain collision-free flight and formation keeping in the presence of wind and sensor noise, although performance remains highly sensitive to the tuning of interaction strengths and the management of delays \cite{vasarhelyi2014,vasarhelyi2018,zhou2022science}. Data fusion from various sources, including vision and distance measurements \cite{Xu2022}, can increase the update rate of neighbors' state estimation compared to pure communication, but scalability is limited. Complementary evidence from fixed-wing aerial swarms further reveals that flock coherence can break down abruptly when communication range and maximum turning rate are not jointly tuned, suggesting that real-world UAV collectives may approach sharp behavioral boundaries reminiscent of phase-transition thresholds in which small parameter changes produce qualitatively different collective states \cite{Hauert2011Reynolds}. Other studies have shown that embedding flocking rules in model-predictive or optimization frameworks improves safety and performance in cluttered environments \cite{soria2022,toumieh2024}. In parallel, the push towards biologically grounded perception has led to vision-only approaches in which robots coordinate using dedicated active visual markers \cite{Saska2022UVDAR} or raw visual cues rather than explicit relative positions or communication \cite{Saska2024,mezey2025}. Within this context, a central scientific question remains unresolved, namely how variations in the strength of local interaction rules influence the global collective state of an autonomous drone swarm and how these collective states in turn determine the swarm’s ability to respond to external disturbances or threats.

Previous work in computer science, biology, and physics has demonstrated that collective motion can emerge from simple local interaction rules~\cite{Reynolds1987,vicsek1995,Couzin2002}. However, these studies also reveal important limitations that remain insufficiently addressed. In swarm robotics, many flocking models successfully reproduce coordinated motion but often rely on idealized assumptions that neglect key constraints such as communication delays, sensing noise, and motion dynamics~\cite{vasarhelyi2018,Hauert2011Reynolds}. As a result, their predictions do not always translate to real-world systems, where coherence can break down abruptly when these constraints are not properly accounted for~\cite{vasarhelyi2014}. In particular, previous experimental studies have shown that factors such as communication range and turning limitations can critically affect the stability of aerial swarms, although these effects have not been systematically analyzed in terms of phase transitions or critical phenomena~\cite{Hauert2011Reynolds}.

In physics, transitions between disordered and ordered motion are often framed as collective phase transitions. Collective states or phases are characterized by order parameters, which quantify the order (or its absence) in the system, such as the polarization, i.e., the mean orientation of spins in a magnetic material or of fish in a school. At the transition between two collective phases (for instance, induced by a change in interaction parameters between spins or fish), the system is in a \textit{critical state}, which is characterized by large \textit{fluctuations} (variance) of the order parameter and multistability. In addition, the fluctuation-dissipation theorem states that these fluctuations are proportional to the system’s response (i.e., the change in the order parameter) to a perturbation (a small magnetic field for magnetic spins; a small fraction of perturbing fish for a school), with this response quantified by the \textit{susceptibility}. 
This phenomenology has also been shown to hold in realistic behavioral models for fish schools~\cite{calovi2015collective,Lin2026}.
As a consequence, the divergence of fluctuations at a critical point coincides with maximal sensitivity of the system to perturbations. From a biological perspective, it can be beneficial for a group of animals (e.g., fish) to operate near a critical point, which offers maximal sensitivity and reactivity to perturbations in their environment, enabling efficient large-scale reorganizations \cite{Stanley1971,munoz2018colloquium}. 

At the same time, empirical studies of animal groups reveal that collective motion is intrinsically dynamic and involves continuous transitions between different behavioral states~\cite{Lopez2012,Calovi2014,Wang2022impact,Wang2025}. These dynamics are accompanied by high fluctuations in global order parameters such as polarization, which are increasingly interpreted as signatures of proximity to critical regimes where information transfer and responsiveness to environmental changes are maximized~\cite{calovi2015collective,Lin2026,Cavagna2010,Bialek2014,Attanasi2014Information,gomez2023fish,Puy2024,Lin2025}. 

Despite these advances, a clear gap remains between theoretical models, biological observations, and robotic implementations. While prior work has shown that adjusting interaction parameters can modulate cohesion and stability~\cite{vasarhelyi2018,Couzin2002,Wang2022impact,Wang2025}, it remains unclear how such changes reorganize collective behavior at a deeper level, or whether they give rise to distinct dynamical regimes with specific functional properties. In contrast, the existence and functional implications of these phase transitions remain largely unexplored in autonomous aerial swarms~\cite{Dorigo2021SwarmRobotics}. It remains unclear whether UAV swarms exhibit sharp and reproducible transitions between distinct collective states under realistic sensing, communication, and dynamical conditions. This raises a central question of whether operating near such transition regimes can enhance responsiveness and resilience to external perturbations, such as intrusions or environmental disturbances.

Recent experimental and theoretical studies have begun to explore this question. It has been shown that programmable robot swarms can exhibit maximal responsiveness near an order–disorder transition, provided that this transition is driven by alignment interactions~\cite{Lei2023}. Other work has demonstrated that microrobotic swarms are able to mount effective collective responses to external threats even when individual agents have only limited information, highlighting the robustness of decentralized coordination~\cite{ChenBechinger2022}. Additional studies suggest that the functional benefits associated with criticality may depend not only on responsiveness but also on the spatial organization of the group.~\cite{KlamserRomanczuk2021}. These studies provide important insights but leave open key questions regarding the emergence of criticality in aerial swarms operating under realistic constraints and the role of simple interaction rules in shaping both responsiveness and robustness.

\begin{figure}[ht]
 \centering
 \includegraphics[width=0.99\textwidth]{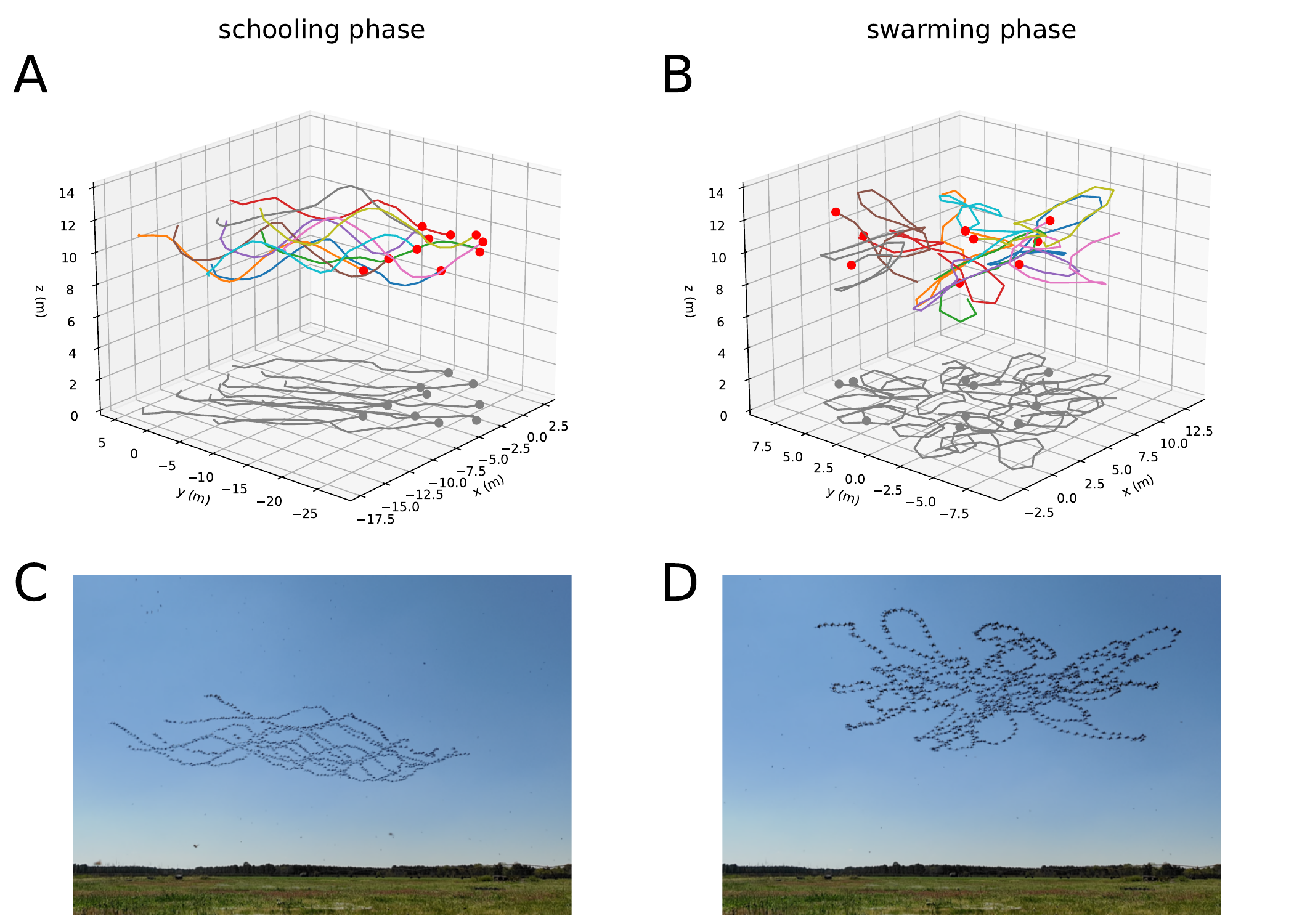}
    \caption{\textbf{Representative collective-motion patterns of the drone swarm in experimental flights.} (\textbf{A}) Schooling: a highly aligned and cohesive collective state. (\textbf{B}) Swarming: a compact but weakly aligned collective state. (\textbf{C}) and (\textbf{D}) Ground-based views of the drone swarm in flight, shown as composite images built from 20 seconds of video recorded at 4 fps, corresponding respectively to the schooling (\textbf{A}) and swarming (\textbf{B}) phases.}
    \label{fig:collective_patterns}
\end{figure}

The present study addresses this gap by combining large-scale experiments with a bio-inspired 3D flocking model specifically designed to reveal phase-transition dynamics in drone swarms. The interaction rules implemented in this model are grounded in quantitative studies of animal collective behavior, which have shown that coordinated motion in fish schools emerges from the combined effects of attraction and alignment responses to neighbors~\cite{calovi2018disentangling, lei2020computational,Wang2022impact}. These interaction components correspond to measurable behavioral processes that govern how individuals adjust their heading and spatial position in response to social cues. In this framework, we use a minimal local-interaction framework in which each drone modulates its motion based on distance and heading information from a small set of influential neighbors (two in this case), controlled by only two gains that quantify alignment and attraction~\cite{verdoucq_bio-inspired_2023}. By systematically varying these gains, we map a phase diagram that spans \textit{(i)}~swarming --- a collective pattern defined by loose aggregation, with relatively strong attraction between agents but low alignment (low polarization; Figure~\ref{fig:collective_patterns}B-D);  \textit{(ii)}~schooling --- a highly coordinated collective motion exhibiting strong alignment (high polarization; Figure~\ref{fig:collective_patterns}A-C); and \textit{(iii)}~an intermediate critical region where polarization fluctuations and reorganization capacity (susceptibility) are maximal. Using both numerical simulations and outdoor free-flight experiments with a swarm of UAVs, we quantify how polarization, dispersion, and minimal inter-agent distance evolve across these regimes and how they are modulated by internal parameter changes and external perturbations, including targeted intrusions by an external drone referred to as the intruder.

Our specific objectives are threefold. First, we seek to demonstrate experimentally that an autonomous drone swarm can undergo a clear transition between collective phases under real-world conditions and to identify the associated critical region in terms of interaction strengths. Second, we investigate the dynamics and asymmetry of switching between schooling and swarming states when alignment strength is abruptly increased or decreased, and assess the implications of these transition times for mission design. Third, we aim to determine how the location of the swarm in the phase diagram influences its responsiveness and resilience to threats by measuring the susceptibility and recovery time when the group is perturbed by an intruder approaching from different directions.

Taken together, these objectives allow us to address the broader question of how simple interaction gains determine where an autonomous aerial swarm lies on the continuum between stability and responsiveness, and whether operating near the critical transition region can improve the swarm’s ability to cope with disturbances without compromising safety. More generally, this perspective recasts flocking not merely as a behavior to replicate but as a controllable collective state whose collective properties can be shaped directly through gain modulation. This framing opens the possibility of using adaptive interaction gains as a form of collective gain control, enabling UAV swarms to collectively and autonomously shift between efficient, stable transit modes, and highly responsive configurations suited for exploration, rapid reorientation, or defensive maneuvers. In the following sections, we describe how we implement this framework on a real drone swarm, how the resulting collective states emerge across interaction regimes, and how these states govern the swarm’s response to both internal and external perturbations in realistic outdoor conditions.

\section*{Results}

\subsection*{Collective states and criticality}

We first present how the bio-inspired interaction rules shape the collective dynamics of the drone swarm across baseline, perturbed, and transition conditions, combining experimental observations with simulation results.

To investigate how bio-inspired interaction principles shape swarm dynamics, we developed an experimental framework in which each quadrotor UAV continuously adjusts its motion in response to interactions with a limited set of \emph{influential} neighbors. These interactions are governed by two control parameters, the alignment gain $\gamma_{\rm Ali}$ and the attraction gain $\gamma_{\rm Att}$, which directly correspond to the two main components of social interactions identified in fish schools~\cite{calovi2018disentangling, lei2020computational}. The alignment gain quantifies the tendency of an individual to align its heading with that of its neighbors, while the attraction gain regulates inter-individual spacing and promotes group cohesion. Both gains act on the variations in the drone’s course direction, consistent with experimental evidence showing that fish primarily control their heading in response to social cues. The resulting control laws are given by:
\begin{align}
\delta {\rm speed} & = \delta {\rm speed}^{\rm social} + \delta {\rm speed}^{\rm nav}, \label{eq:d_speed} \\
\delta {\rm climb} & = \delta {\rm climb}^{\rm social} + \delta {\rm climb}^{\rm nav} + \delta {\rm climb}^{\rm intruder}, \label{eq:d_climb}\\
\delta {\rm course} & = \delta {\rm course}^{\rm social}(\gamma_{\rm Ali}, \gamma_{\rm Att}) + \delta {\rm course}^{\rm nav} + \delta {\rm course}^{\rm intruder}, \label{eq:d_course}
\end{align}
where $\left[ \delta {\rm speed}, \delta {\rm climb}, \delta {\rm course} \right]$ are the variations of horizontal speed, climb rate (vertical speed), and course (horizontal direction of motion relative to the ground), respectively. They are a combination of the social interactions between agents (${\rm social}$), the navigation objectives (${\rm nav}$), and the reaction to external intruders (${\rm intruder}$).

\begin{figure}[ht]
 \includegraphics[width=1.\linewidth]{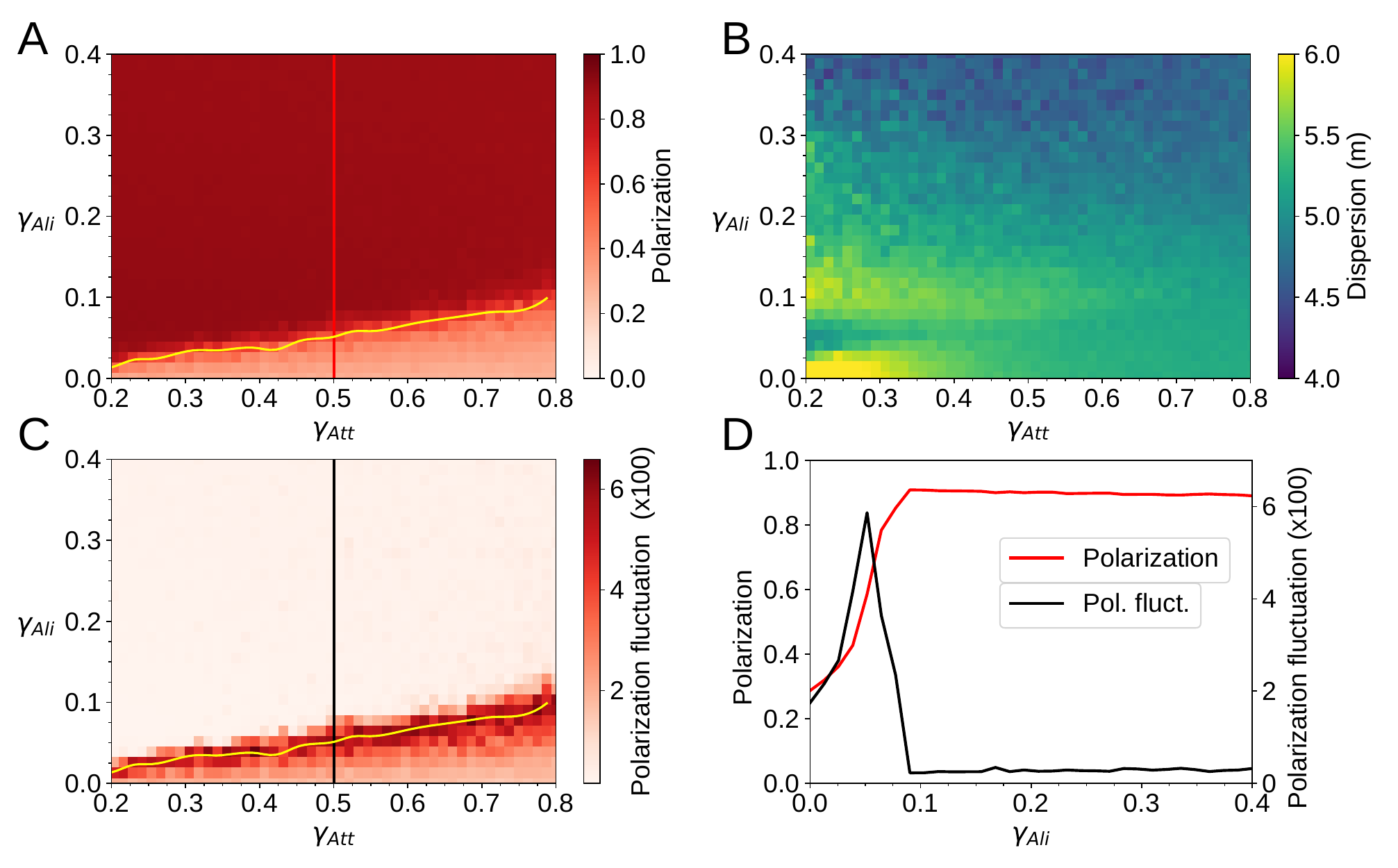}
    \caption{\textbf{Phase-diagram structure of collective motion under varying attraction and alignment strength}. Heatmaps show the average (\textbf{A}) polarization, (\textbf{B}) dispersion, and
    (\textbf{C})~polarization fluctuations, together with (\textbf{D}) the transects at $\gamma_{\rm Att}=0.5$, which are the focus of this study. For each parameter pair $(\gamma_{\rm Ali}, \gamma_{\rm Att})$, we ran ten 300-s simulations with a swarm of 10 drones and report the means across runs.}
    \label{fig:phase-diagram}
\end{figure}

Using only two control parameters, the alignment and attraction intensity, $\gamma_{\rm Ali}$ and $\gamma_{\rm Att}$, both acting on the course variation from social interactions, we were able to generate a broad repertoire of collective states, ranging from highly aligned schooling to compact but unpolarized swarming (see Figure~\ref{fig:phase-diagram} for $N=10$ and Supplementary~Figure~1 for $N=20$). To establish the trade-off between these two key parameters, we built a phase diagram by varying $\gamma_{\rm Ali}$ and $\gamma_{\rm Att}$ in a range [0--0.4] and [0.2--0.8], respectively, consisting of ten simulation runs per grid point. These ranges for the attraction and alignment strength have been empirically determined, where lower values lead to weak interactions and a loss of cohesion, while larger values tend to saturate the command inputs, which can lead to unwanted behaviors, such as oscillations. To quantify the group-level dynamics, we used standard indicators widely employed in the study of natural collective motion, namely polarization, dispersion, and minimal inter-agent distance~\cite{lei2020computational,Wang2022impact,Lin2025}.
A detailed description of the computational model summarized by equations \ref{eq:d_speed} to \ref{eq:d_course} and the metrics used to quantify collective states is provided in Methods. 

The resulting phase diagram reveals a clear phase transition between the swarming and schooling states (see Figure~\ref{fig:phase-diagram}A-C), highlighting the so-called critical region. The dispersion is considerably higher (see Figure~\ref{fig:phase-diagram}B) for the lowest levels of interaction gains, since without enough alignment and attraction forces, the drones are flying without group cohesion (low attraction) and will mostly move forward in their own direction (low alignment).

The fluctuations of the polarization are maximal along the critical line separating the swarming and schooling phases (see Figure~\ref{fig:phase-diagram}C) and their magnitude along this line increases with group size. Indeed, in Supplementary~Figure~1C (for $N=20$), the region of maximal fluctuations is even more clearly identified than in Figure~\ref{fig:phase-diagram}C (for $N=10$), illustrating the fact that the transition becomes sharper as the system size increases. 
In fact, identifying the region of maximal fluctuations of the order parameter (here, the polarization) in a phase diagram is an alternative way to locate the critical regions~\cite{calovi2015collective,Lin2025,Lin2026}.
Moreover, and quite generally, the response of a system to an external perturbation is proportional to the fluctuations of the order parameter in the absence of perturbation, with a proportionality coefficient called the susceptibility. This property is reminiscent of the fluctuation-dissipation theorem in physics and has also been shown to apply to models of interacting self-propelled particles in the context of fish school models~\cite{calovi2015collective,Lin2025,Lin2026}. Hence, the intrinsic fluctuations of a system also provide insight into its sensitivity and its ability to respond to a perturbation (susceptibility).

Doubling the swarm size increases the slope of the curve defining the critical region by a factor of two (see Figure~\ref{fig:phase-diagram}A and Supplementary Figure~1A) and produces a corresponding two-fold shift in both the peak of the polarization fluctuations and the saturation point of the polarization curve (see Figure~\ref{fig:phase-diagram}D and Supplementary Figure~1D). These changes indicate that a substantially stronger alignment force is required to reach the same level of polarization in the larger swarm, which in turn results in a marked increase in dispersion at low alignment gains (see Figure~\ref{fig:phase-diagram}B and Supplementary Figure~1B).

To constrain the parameter space for our outdoor experiments, we focused on a transect of the phase diagram (see Figure~\ref{fig:phase-diagram}D), defined by a fixed attraction strength ($\gamma_{\rm Att} = 0.5$) and systematically varying alignment gain $\gamma_{\rm Ali}$ between $0.025$ and $0.4$. This choice allowed us to focus on a specific transect of the phase diagram that contains well-defined swarming and schooling regimes (see Figure~\ref{fig:collective_patterns}), as well as a critical region between them for $\gamma_{\rm Ali} \in [0.1, 0.2]$.
Figure~\ref{fig:phase-diagram}D confirms that the transition from a low to high polarization coincides with a maximum of the fluctuations of the polarization, as is already apparent in the overall phase diagram (see Figure~\ref{fig:phase-diagram}A and C). Again, and in accordance with the general relation between the fluctuations of the order parameter and the susceptibility to a perturbation, a drone swarm near the critical region should have a maximal sensitivity to a change in its environment.
For trials involving intruder perturbations, indicators were computed on the time interval during which the intruder approached the swarm barycenter from a distance of 25\,m, enabling a direct assessment of the swarm’s response to an imminent disturbance.

\begin{figure}[tbp]
  \centering
  \vspace{-1em}
  \includegraphics[width=.73\textwidth]{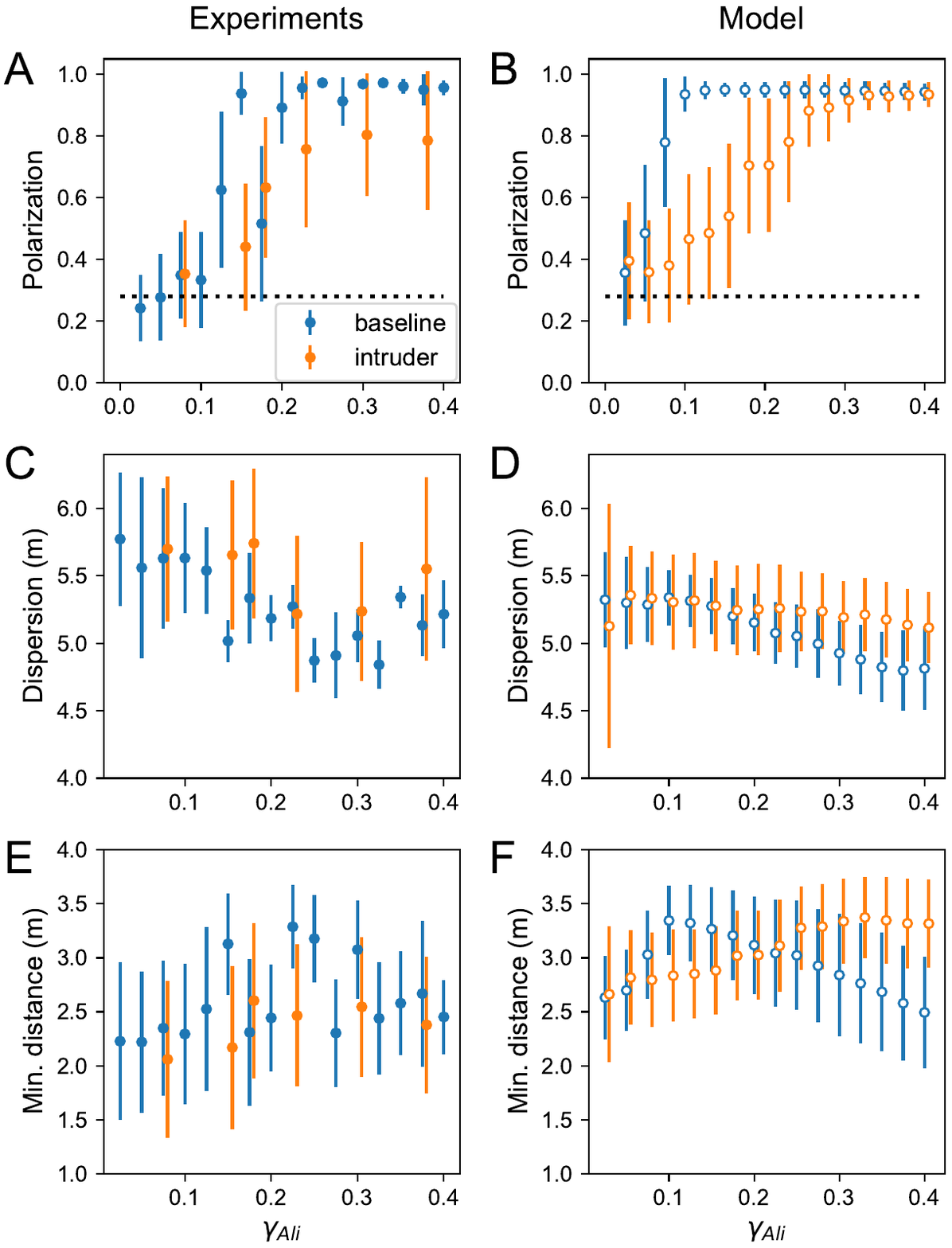}
  \vspace{-2em}
  \caption{\textbf{Effects of alignment strength on swarm polarization, dispersion, and inter-agent spacing.} Swarm polarization (\textbf{A, B}), dispersion (\textbf{C, D}), and minimum inter-agent distance (\textbf{E, F}) measured in three-dimensional space. (\textbf{A, C, E}) Outdoor field experiments performed with Parrot Bebop 2 UAVs. (\textbf{B, D, F}) Corresponding numerical simulations. All experiments used $\gamma_{\rm Att} = 0.5$, while $\gamma_{\rm Ali}$ varied between $0.025$ and $0.4$. The baseline condition consisted of a swarm of ten drones, whereas the perturbation condition included an additional intruder drone. Simulations comprised 100 independent runs for each value of $\gamma_{\rm Ali}$ each lasting 5 min. In the outdoor experiments, trial durations varied among $\gamma_{\rm Ali}$ values. Symbols represent mean values and error bars indicate standard deviations.. The horizontal dashed line in panels (\textbf{A,B}) indicates the expected polarization of a randomly oriented group equal to % $1/\sqrt{N} = 0.32$ or
  $0.5 \sqrt{\pi/N} = 0.28$ for $N=10$.}
\label{fig:error_bars}
\end{figure}

With these parameter settings, we performed a series of simulations consisting of 100 runs for each $\gamma_{\rm Ali}$ value, with a fixed duration of five minutes per value, whereas in the outdoor experiments the duration for each $\gamma_{\rm Ali}$ value varied due to experimental contingencies.
The baseline condition consisted of a swarm of ten cooperative drones, and the test condition introduced an additional intruder drone. The aggregated results of both field experiments with Parrot Bebop 2 UAVs and numerical simulations presenting the means and the standard deviations of swarm polarization, dispersion, and minimal inter-agent distance are shown in Figure~\ref{fig:error_bars}.

In the baseline condition in simulation, without an intruder, polarization saturates at $\gamma_{\rm Ali} = 0.1$ and exhibits a broad variance peak between $0.05$ and $0.075$, corresponding to the critical region, consistent with the expected increased susceptibility (Figure~\ref{fig:error_bars}B). Under intruder perturbation, the variance increases well beyond the baseline critical region, and the saturation shifts toward higher $\gamma_{\rm Ali}$ values (around $0.3$), highlighting the swarm’s ability to retain high susceptibility in the face of external disturbances. Field measurements (Figure~\ref{fig:error_bars}A) follow a similar trend in terms of magnitude of variations, with some noticeable differences. First,  the empirical critical region is shifted and goes from $\gamma_{\rm Ali} = 0.1$ to $0.2$. Second, the polarization drop due to the intruder is visible but with a lower intensity (for $\gamma_{\rm Ali} \in [0.1, 0.2]$ compared to $[0.5, 2.5]$ in simulation). Finally, an outlier at $\gamma_{\rm Ali} = 0.175$ disrupts the trend, which is plausibly explained by limited sampling and proximity to the critical region. This simulation-to-reality gap has several explanations. The horizontal shift is most likely attributable to the limitations of the drone model, which was simplified to a first-order model (see equation~\ref{eqn:fish_model} in Methods). The outlier in the critical region is consequence of the limited number of experimental data available. In general, the higher noise is the result of the communication latency, the transmission errors, and the disturbances from wind and gust that are not present in the simulation.

Swarm dispersion (Figure~\ref{fig:error_bars}D) continuously decreases in simulation with increasing $\gamma_{\rm Ali}$ as expected, since coherent schooling is spatially more compact than swarming. Baseline and perturbation trends are similar within the critical region, but diverge markedly in the schooling state, where the intruder induces a strong repulsive expansion of the swarm. Around the critical region, this effect diminishes because drones naturally maintain larger separations. Although the empirical values (Figure~\ref{fig:error_bars}C) are slightly higher, dispersion patterns are consistent between experiments and simulations until $\gamma_{\rm Ali} = 0.3$ where it is slightly increasing again. There is no evidence from the data or associated videos that an abnormal external condition would have impacted the observed dispersion. This discrepancy could be related to the natural perturbations and the limited dataset and associated variance.

Minimal inter-agent distance (Figure~\ref{fig:error_bars}E-F) peaks in the critical region, again reflecting maximal susceptibility and rapid reorganization capacity. It then decreases monotonically toward both the schooling and swarming regimes. The intruder strongly reduces the minimal distance within the critical region by destabilizing the swarm, whereas in the schooling state it causes a compensatory increase of approximately 1\,m as the group briefly expands to accommodate the passing intruder. Experimental values are within the same range, but due to strong noise, the overall trend is difficult to assess.

\subsection*{Baseline dynamics near the critical region}

\begin{figure}
    \centering
    \includegraphics[width=1.1\linewidth]{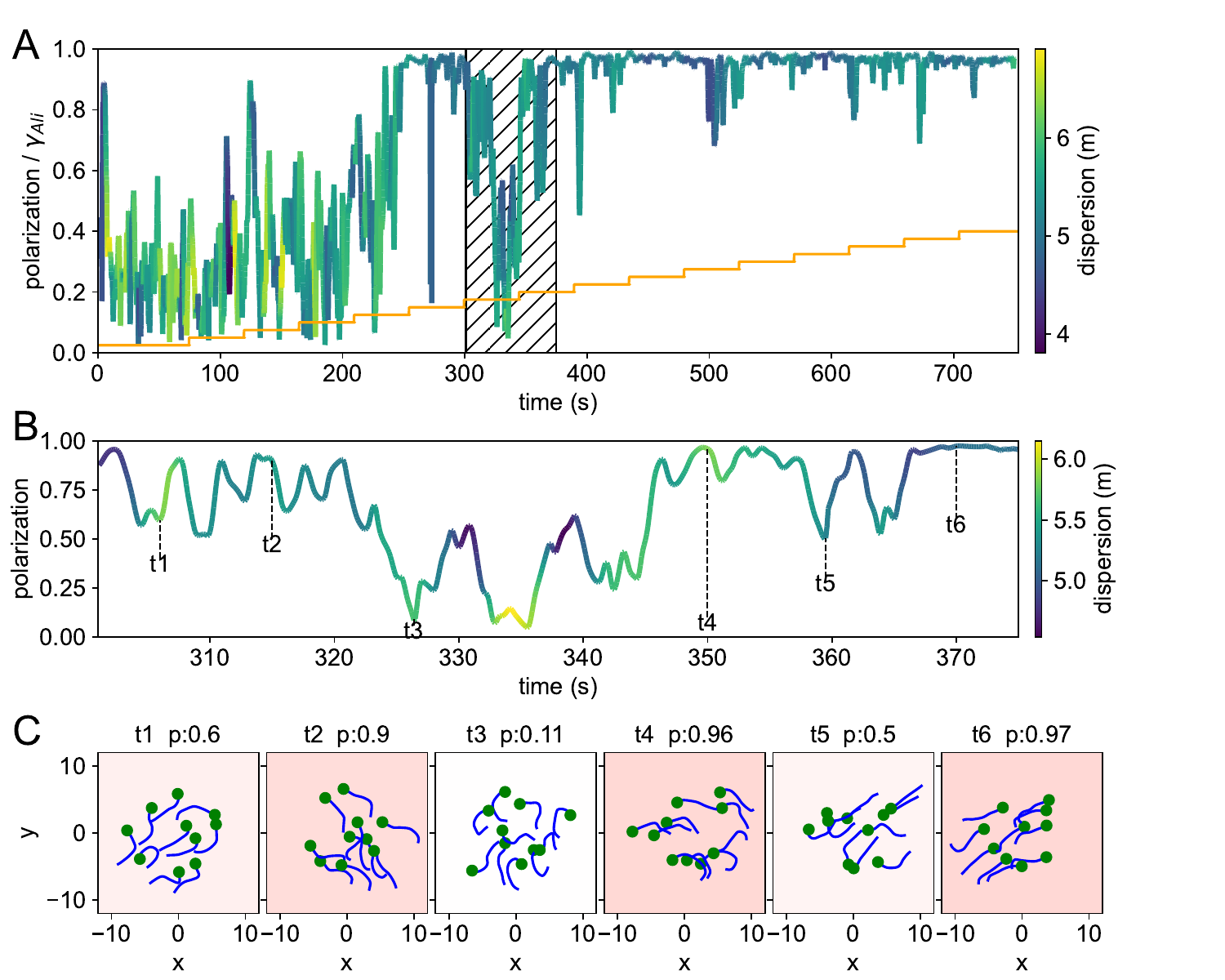}
    \caption{\textbf{
    Baseline Dynamics of a ten-UAV swarm near the critical region during the outdoor field experiment}. (\textbf{A}) Time evolution of polarization and dispersion, color-coded by metric. The orange curve indicates the time-dependent variations of $\gamma_{\rm Ali}$ during this run. (\textbf{B}) Enlarged view of a part of the critical region ($\gamma_{\rm Att}=0.5$; $\gamma_{\rm Ali}=0.175$ and $\gamma_{\rm Ali}=0.2$), showing six selected timestamps. (\textbf{C}) Corresponding swarm configurations in the $X$-$Y$ plane, where points represent the drones’ instantaneous positions and tails indicate their trajectories over the preceding five seconds, and background shading reflects polarization level (see Supplementary Movie~1).}
    \label{fig:time_series_baseline}
\end{figure}

To characterize the swarm’s temporal response in the critical region, Figure~\ref{fig:time_series_baseline}A shows the evolution of polarization and dispersion across consecutive increases of $\gamma_{\rm Ali}$ (the orange step curve). For low $\gamma_{\rm Ali}$, the swarm exhibits low, highly variable polarization (swarming), whereas higher $\gamma_{\rm Ali}$ produces high and stable polarization (schooling).
A magnified segment of the time series, part of the critical region, is shown in Figure~\ref{fig:time_series_baseline}B for $\gamma_{\rm Ali} = 0.175$ and $\gamma_{\rm Ali} = 0.2$, with corresponding 2D swarm configurations displayed at six annotated timestamps in Figure~\ref{fig:time_series_baseline}C. The inset reveals that an abrupt increase in $\gamma_{\rm Ali}$ initially destabilizes the swarm: polarization decreases sharply up to $t1$ while the swarm still attempts to maintain schooling. As the system enters the schooling regime ($t2$), polarization decreases further ($t3$) before recovering at $t4$, after $\gamma_{\rm Ali}$ has increased again. A second, smaller polarization drop occurs at $t5$, consistent with the heightened susceptibility characteristic of the critical region (see Supplementary Movie~5).

\subsection*{Dynamics of switching between schooling and swarming}

\begin{figure}
    \centering
    \includegraphics[width=1.\linewidth]{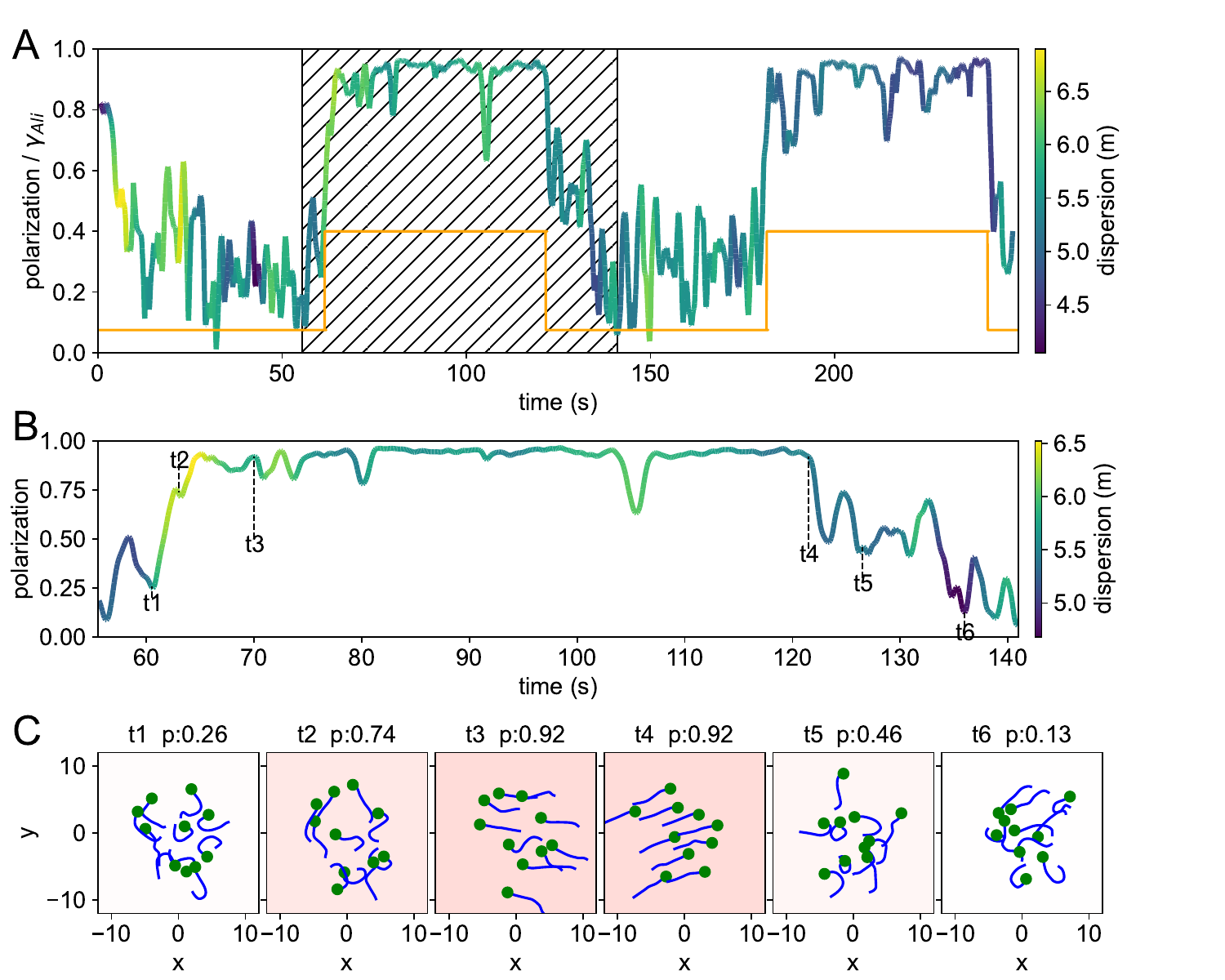}
    \caption{\textbf{
    Swarm dynamics across repeated phase transitions between swarming and schooling during the outdoor field experiment}. The ten-UAV swarm was tested under baseline conditions while the alignment strength $\gamma_{\rm Ali}$ was alternated between low (0.075) and high (0.4) values for $\gamma_{\rm Att}=0.5$, corresponding respectively to swarming and schooling behaviors. (\textbf{A}) Time evolution of polarization and dispersion, color-coded by metric. The orange curve indicates the time-dependent variations of $\gamma_{\rm Ali}$ throughout the run. (\textbf{B}) Enlarged view of a full switching cycle (swarming → schooling → swarming), highlighting six selected timestamps. (\textbf{C}) Corresponding $X$-$Y$ swarm configurations, where points represent the drones’ instantaneous positions and tails trace their trajectories over the previous five seconds, and background shading reflects polarization level (see Supplementary Movie~2).}
    \label{fig:time_series_phase_switch}
\end{figure}

To probe the swarm’s dynamical response to abrupt internal perturbations, we conducted a separate experiment in which $\gamma_{\rm Ali}$ was repeatedly switched between $0.075$ (swarming) and $0.4$ (schooling) (Figure~\ref{fig:time_series_phase_switch}). The inset (Figure~\ref{fig:time_series_phase_switch}B-C) shows that transitions from swarming to schooling occur within approximately 5\,s (between $t1$ and $t2$), whereas the reverse transition takes about 15\,s (between $t4$ and $t6$). This three-fold asymmetry is consistent across trials. A plausible explanation is that high $\gamma_{\rm Ali}$ actively enforces rapid alignment, whereas loss of alignment at low $\gamma_{\rm Ali}$ is not forced but emerges gradually through accumulated perturbations (see Supplementary Movie~6).
Also note that the two situations from Figure~\ref{fig:phase-diagram}, schooling and swarming, are extracted from the data in Figure~\ref{fig:time_series_phase_switch}, starting at times 142 and 194\,s, respectively.

\subsection*{Swarm response to an approaching intruder}

\begin{figure}
    \centering
    \includegraphics[width=1.\linewidth]{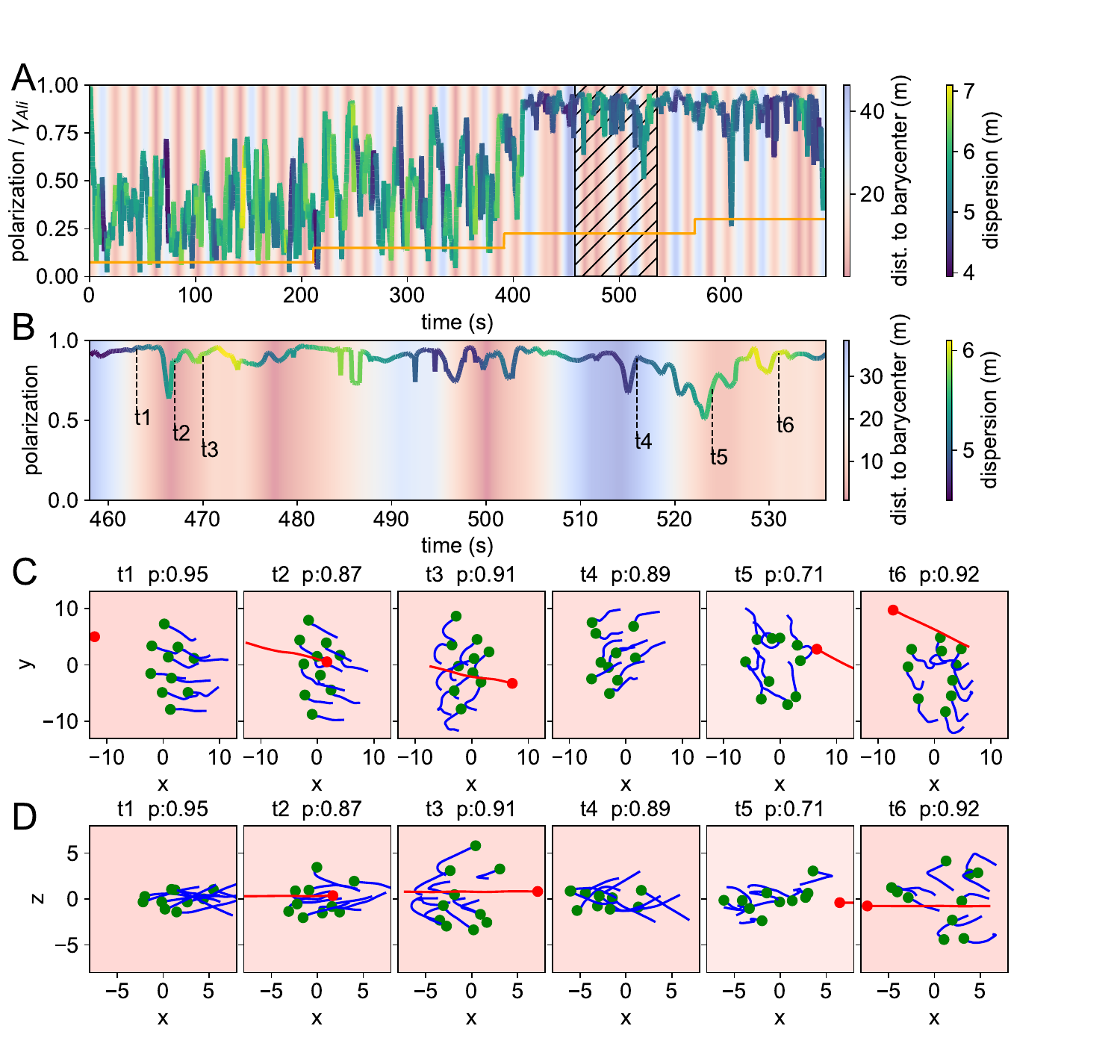}
    \caption{\textbf{
    Swarm response to an intruder near the critical region during the outdoor field experiment}.
    (\textbf{A}) Time evolution of polarization and dispersion (line colors) together with the distance between the intruder and the swarm barycenter (background shading). 
    The orange curve shows the temporal variations of $\gamma_{\rm Ali}$ [0.075, 0.15,  0.225, 0.3] throughout the run; the hatched area marks the region shown in B.
    (\textbf{B}) Enlarged view of the hatched region (corresponding to $\gamma_{\rm Att}=0.5$, $\gamma_{\rm Ali}=0.225$, right after the critical region), highlighting six selected timestamps.
    (\textbf{C, D}) Corresponding swarm configurations in the $X$-$Y$ and $X$-$Z$ planes, where points denote the drones' instantaneous positions and tails trace their trajectories over the preceding five seconds, and background shading reflects polarization level (see Supplementary Movie~3).}
    \label{fig:time_series_intruder}
\end{figure}

To evaluate how external agents disrupt swarm coordination, we conducted outdoor field experiments in which a single intruder drone performed 3D flight maneuvers directed toward the swarm barycenter. The intruder maintained a fixed heading once within a predefined proximity and did not interact socially with the swarm.
As noted above, the intruder has a pronounced destabilizing effect on collective dynamics. A representative run is shown in Figure~\ref{fig:time_series_intruder}, in which $\gamma_{\rm Ali}$ varied between $0.075$ and $0.3$. Polarization levels (Figure~\ref{fig:time_series_intruder}A) indicate that the perturbation keeps the swarm in a swarming-like state at higher $\gamma_{\rm Ali}$ values than observed under baseline conditions.
A detailed view near the critical region at $\gamma_{\rm Ali} = 0.225$ is shown in Figure~\ref{fig:time_series_intruder}B,C, where the polarization drop is expected to be the strongest. When the intruder approaches frontally ($t1$–$t3$), the swarm performs a stereotypical collective turn (see also Figure~\ref{fig:intruder_sim}C). This rapid reorientation causes a sharp, transient drop in polarization. When the intruder approaches from the side ($t4$–$t6$), polarization decreases during the encounter and increases again once the intruder crossed the swarm barycenter. In both situations, the swarm recovers within 5–6\,s and re-establishes high alignment. This heightened responsiveness is characteristic of the critical region and disappears for $\gamma_{\rm Ali} > 0.4$, where the swarm maintains high polarization even while avoiding the intruder (see Supplementary Movie~7).

\begin{figure}
    \centering
    \includegraphics[width=1.\linewidth]{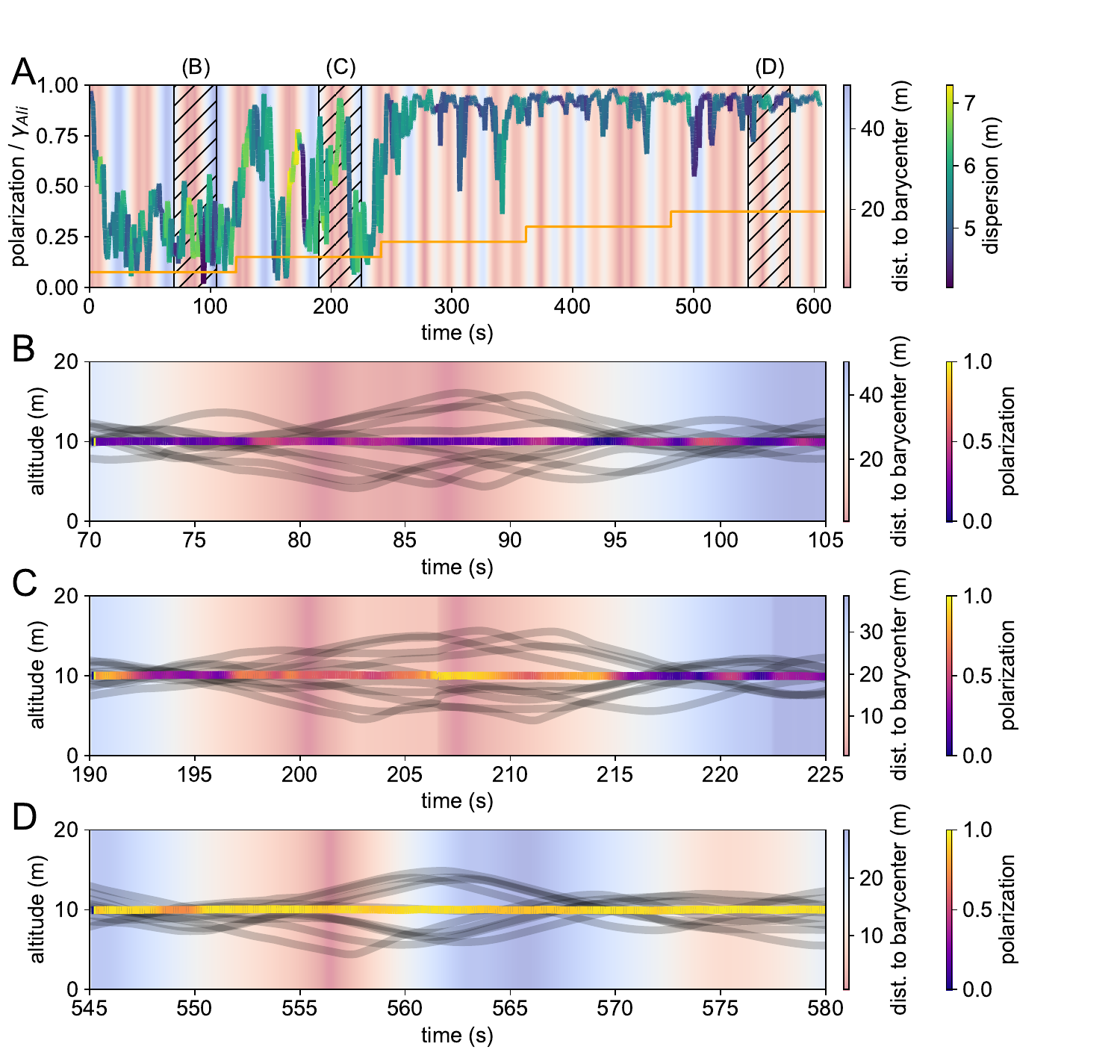}
    \caption{\textbf{
    Swarm response to an intruder in the swarming (\textbf{B}), critical (\textbf{C}) and schooling (\textbf{D}) states, during the outdoor field experiment}. 
    (\textbf{A}) Time evolution of polarization and dispersion (line colors) together with the distance between the intruder and the swarm barycenter (background shading). 
    The orange curve shows the temporal variations of $\gamma_{\rm Ali}$ [0.075, 0.15,  0.225, 0.3, 0.375] throughout the run; the hatched areas mark the swarming, critical, and schooling regions.
    (\textbf{B, C, D}) Time evolution of the altitude in the hatched regions of the swarm (in black) and the intruder (colored line at constant 10\,m, highlighting the swarm polarization) together with the distance between the intruder and the swarm barycenter (background shading) (see Supplementary Movie~4).
}
    \label{fig:alt_time_series_intruder}
\end{figure}

\begin{figure}[ht]
    \centering
    \includegraphics[width=1.\linewidth]{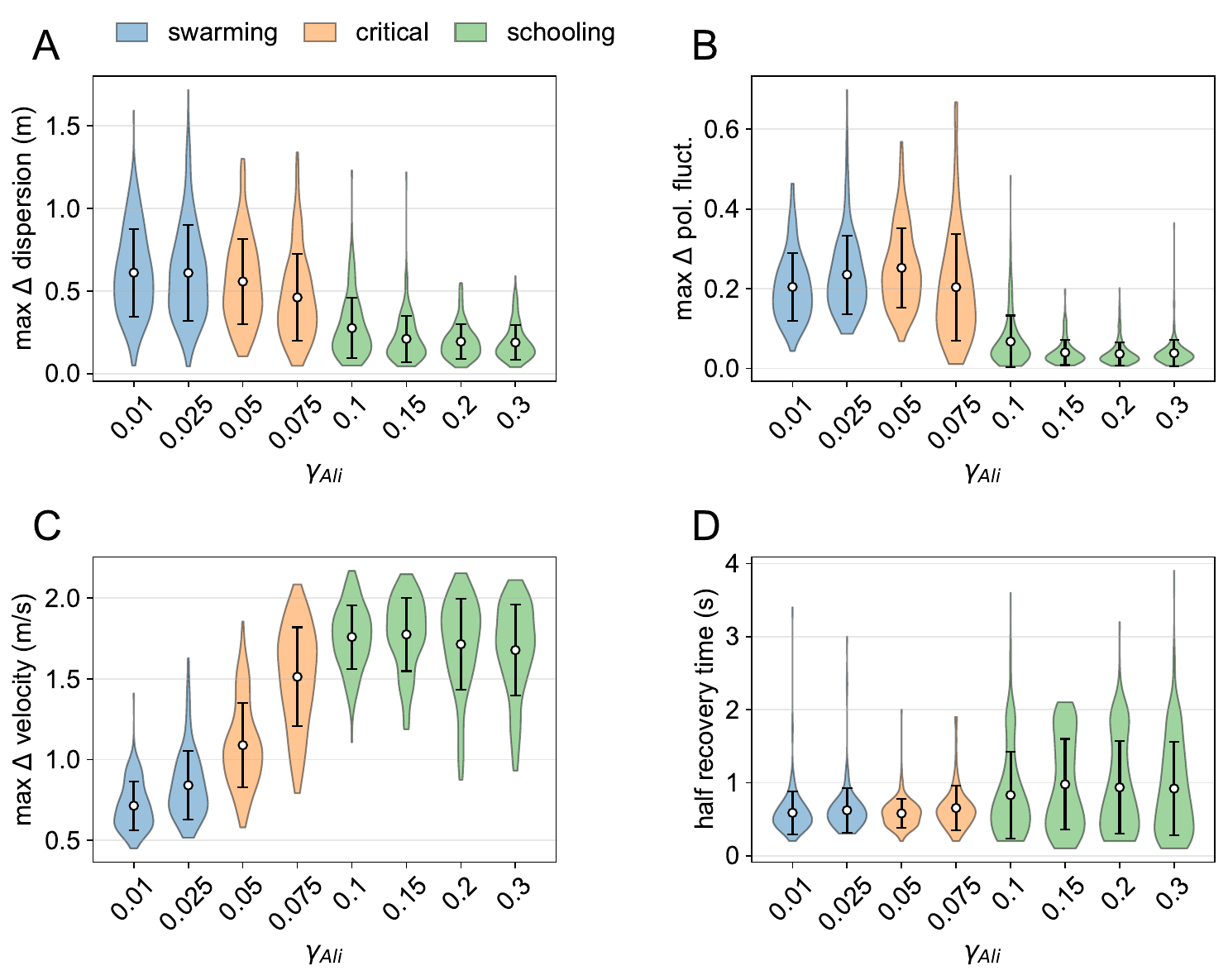}
    \caption{
\textbf{Characterization of swarm responsiveness and resilience under intruder frontal attack.} Simulations with intruder angle of attack of $0^\circ$. (\textbf{A}) maximal transient swarm opening -- peak increase in dispersion relative to baseline; (\textbf{B}) maximal amplification of polarization fluctuations; (\textbf{C}) maximal swarm velocity change; and (\textbf{D}) half recovery time following perturbation. Critical swarms ($\gamma_{\mathrm{Ali}} \in [0.05, 0.075]$) exhibit substantially larger transient opening and polarization-fluctuation amplification, indicating stronger collective responsiveness to the approaching intruder and recover substantially faster after perturbation while requiring smaller global heading reorientation, suggesting that threat mitigation is achieved through adaptive local restructuring rather than large-scale rigid turning.
}
    \label{fig:intruder_sim}
\end{figure}

Another representative example of swarm response to the approaching intruder is shown in 
Figure~\ref{fig:alt_time_series_intruder}. We have magnified the altitude trajectories of all drones during three periods of within a single experimental run, highlighted in Figure~\ref{fig:alt_time_series_intruder}A, corresponding to swarming (Figure~\ref{fig:alt_time_series_intruder}B), critical (Figure~\ref{fig:alt_time_series_intruder}C) and schooling (Figure~\ref{fig:alt_time_series_intruder}D) states. In all cases (B-C-D) the approaching intruder induced a transient vertical expansion of the swarm that splits into two groups, and rapidly re-establishes its vertical cohesion approximately 5 s after the intruder goes away. However, the collective response differed markedly across dynamical regimes. In the swarming state, vertical splitting occurred while maintaining relatively low polarization. In the schooling state the vertical split was achieved while preserving strong alignment with only limited transient decreases in polarization. By contrast, in the critical state the intruder attack simultaneously induced a strong increase in both vertical dispersion and swarm polarization, revealing a particularly sensitive and adaptive response to external perturbation. 

To further characterize the mechanisms underlying these experimental observations, we extended the intruder experiments using numerical simulations comprising 200 repetitions for each condition and multiple intruder attack directions spanning from $0^\circ$ to $315^\circ$. Collective responses remained qualitatively similar across attack geometries and showed only weak dependence on the direction of approach. For clarity, we therefore present the representative $0^\circ$ condition in Figure~\ref{fig:intruder_sim} (for reference see Supplementary Figures 2 and 3). Rather than relying exclusively on susceptibility measurements, we quantified the swarm response using four complementary observables capturing distinct aspects of collective adaptation: transient swarm opening, amplification of polarization fluctuations, swarm velocity change, and post-perturbation recovery time.

The swarming regime exhibits a maximum variation of dispersion that is, on average, approximately twice as large as that observed in the schooling regime (Figure~\ref{fig:intruder_sim}A), while the corresponding change in swarm velocity is roughly twofold smaller (Figure~\ref{fig:intruder_sim}C). These results suggest that weakly aligned groups (swarming phase) respond to an approaching intruder primarily through spatial reorganization, expanding the swarm while maintaining a relatively stable collective velocity. By contrast, highly aligned groups (schooling phase) remain cohesive and strongly polarized during the perturbation (Figure~\ref{fig:intruder_sim}C), but respond by modifying their collective velocity and direction of motion to avoid the intruder (Figure~\ref{fig:intruder_sim}C). The critical regime ($\gamma_{\mathrm{Ali}} \in [0.05, 0.075]$) consistently exhibits a strong transient opening response, allowing drones to locally reorganize and create larger spatial gaps during intruder passage (Figure~\ref{fig:intruder_sim}A), similar to swarming phase. However, the velocity response is about 1.5 stronger than swarming, almost to the level of schooling, indicating a substantial collective reorientation of the group. This interpretation is further supported by the maximum variation of polarization fluctuations observed in the critical region (Figure~\ref{fig:intruder_sim}B), which reflects an enhanced capacity for rapid collective reorganization in response to external perturbations.
Importantly, stronger responses were not associated with slower stabilization; on the contrary, critical swarms recovered significantly faster following a perturbation (Figure~\ref{fig:intruder_sim}D), especially compared to schooling.
Together, the experimental and simulation results demonstrate that swarms in a critical state achieve spatial reorganization while requiring moderate changes in collective speed, indicating that threat mitigation relies primarily on adaptive local restructuring rather than large-scale acceleration of the entire group, balancing responsiveness and robustness.

\section*{Discussion}

In this study, we investigated how a drone swarm can coordinate, adapt, and respond to disturbances through simple bio-inspired local rules. Our goal was to design a 3D flocking algorithm capable of producing multiple collective motion phases and to determine how these phases shape swarm resilience when confronted with a malicious intruder. We sought to identify the interaction regimes that sustain stability, promote flexibility, and support rapid reorganization when the swarm is driven away from equilibrium. To address this challenge, we developed a minimal local-interaction framework based on alignment and attraction, and validated it through both numerical simulations and outdoor free-flight experiments. This experimental strategy enabled systematic modulation of interaction strengths, observation of emergent global patterns, and quantification of the swarm’s real-time response to external perturbations. Taken together, these methods provided a controlled and realistic platform to study the capabilities and limitations of bio-inspired collective motion in three dimensions.

Our results demonstrate that a swarm of drones can express multiple collective motion phases using only two interaction gains, despite limiting the exchange of information to position and heading. The effect of the communication range is not considered here but have been already explored in \cite{verdoucq_bio-inspired_2023}. The drones, relying solely on heading and distance estimates from the two most influential neighbors, displayed sharp transitions between compact swarming and highly aligned schooling, generated by a modified bio-inspired interaction model derived from the mechanisms governing coordinated swimming in rummy nose tetra fish \cite{calovi2018disentangling, lei2020computational,Wang2025}. These dynamics parallel those found in other group-living animals ranging from the fluid collective motion of many other aquatic species \cite{tunstrom2013collective, Kanso2021} to the highly coordinated aerial maneuvers of bird flocks \cite{heppner1997, cavagna2014, cavagna2018}, and reflect the behaviors predicted by classical models of collective motion \cite{vicsek2012}. Comparable patterns have been reported in outdoor robotic swarms using vision-based localization \cite{Saska2022UVDAR,Saska2024,mezey2025} and minimal communication \cite{zhou2022science}, reinforcing the idea that robust coordination emerges from simple local interactions between drones. 

These findings also complement research on confined-environment flocking, where drones maintain cohesion despite obstacles and limited communication \cite{vasarhelyi2018}. On the engineering side, the ability to generate diverse collective states with few parameters reduces sensing requirements, computational load, and communication overhead, supporting the scalability of the approach. Nevertheless, the simplicity of the rules constrains the swarm’s ability to encode complex goals or task-specific strategies, an issue also noted in decentralized coverage and formation-control schemes \cite{nath2022hierarchy, alonsomora2019}. This trade-off suggests that combining minimal models with higher-level decision layers may offer a path toward richer autonomy while preserving scalability.

A central outcome of our study is the identification of a critical region between schooling and swarming where polarization fluctuations are maximal and the swarm displays heightened sensitivity and adaptability (high susceptibility). Within this intermediate regime, small perturbations lead to pronounced changes in polarization and spacing, revealing a state of maximal susceptibility \cite{munoz2018colloquium}. Similar signatures of criticality have been documented in biological collectives and serve as a functional mechanism for fast threat detection and rapid decision propagation \cite{calovi2015collective, gomez2023fish, Puy2024, Lin2025,Lin2026}. These findings align with and extend previous work on criticality in collective systems. In agreement with \cite{Lei2023}, we observe that responsiveness is enhanced near a transition controlled by alignment interactions. However, our study goes beyond previous robotic implementations by demonstrating this effect in a fully three-dimensional aerial system subject to real-world constraints. The responsiveness we observe in real drones parallels that seen in self-organized micro-drone collectives operating in forests and cluttered outdoor environments \cite{zhou2022science}, and echoes the adaptive benefits described in theoretical and empirical studies of flocking under environmental constraints \cite{vasarhelyi2018}. The ability to exploit criticality for rapid reorganization may be particularly useful in naturalistic settings where uncertainty, noise, or adversarial elements are present. However, the same sensitivity also increases the system’s vulnerability to measurement errors and environmental fluctuations. This duality underscores the need to select operating regimes that balance responsiveness with stability depending on mission requirements. Our results emphasize that tuning interaction strengths provides a direct and efficient means to shift the swarm between flexible and stable modes, analogous to the adaptive strategies used in decentralized search-and-rescue collectives \cite{horyna2023autonomous}.

Experiments involving intruder drones further clarify how coordination breaks and reforms under adversarial conditions. When approached by an intruder, the swarm performed abrupt turns, lateral expansions, or evasive maneuvers that resemble escape waves in natural systems \cite{Procaccini2011}. The strongest disturbances occurred when the swarm operated near the critical region, where the susceptibility is highest and small perturbations propagate rapidly through the group. Despite these disruptions, the swarm consistently recovered within seconds, demonstrating resilience rooted in local interaction rules. Similar rapid-recovery behaviors have been observed in underwater robot collectives performing self-organized evasive fountain maneuvers \cite{berlinger2021icra, berlinger2021}, confirming that minimal interaction schemes can support effective avoidance strategies across different physical domains. When alignment levels were high, the swarm maintained polarization and adjusted its trajectory with only minimal internal disorder, in agreement with results from outdoor flocking studies showing that strong alignment stabilizes swarm configurations under wind, turbulence, and environmental noise \cite{vasarhelyi2014}. Conversely, when alignment was weak, the swarm was slower to respond and more prone to fragmentation, reinforcing observations from decentralized GPS-denied swarms where weak coupling increases the risk of drift \cite{saska2016gps}. These findings show that phase-dependent susceptibility plays a crucial role in defining the swarm’s risk profile when facing intruders or obstacles and that dynamic adjustment of interaction strengths can enhance resilience.

Our experiments also reveal that transitions between collective phases are inherently asymmetric. The swarm rapidly forms a schooling state when alignment strength increases, yet returns much more slowly to a flexible swarming state when alignment decreases. This asymmetry suggests that ordered configurations are easier to impose than to disperse and that high alignment fosters strong internal correlations that persist even after the forcing is removed. This phenomenon aligns with theoretical predictions from the Vicsek and Cucker–Smale families of models, where ordered phases form cohesive attractors that slow the return to disordered motion \cite{vicsek1995, cucker2007}. Similar asymmetries have also been observed in distributed formation-control systems, where convergence toward structured formations occurs faster than dissolution, due to consensus-based reinforcement \cite{alonsomora2019}. For real-world robotics, this behavior carries implications for mission design. Rapid organization can be triggered by brief alterations of alignment strength, enabling fast transitions into surveillance or defense modes. However, returning to exploratory or flexible modes requires longer timescales—a consideration also emphasized in dynamic-hierarchy flocking approaches, where transitions between leaders and followers exhibit comparable hysteresis \cite{nath2022hierarchy}. Understanding these asymmetric dynamics allows engineers to design controllers that exploit, rather than fight against, the natural tendencies of multi-agent systems.

Finally, our results provide an alternative perspective on the persistence–responsivity trade-off~\cite{Balazs2020}. Instead of relying on adaptive leadership mechanisms, we show that a comparable balance can emerge through the tuning of simple interaction gains. This finding suggests that complex collective properties can arise from minimal control rules, without the need for explicit hierarchy or additional communication channels.

Taken together, these results illustrate that simple, bio-inspired interaction rules can generate coordinated, versatile, and resilient behaviors in drone swarms in real-world conditions. The emergence of multiple collective phases, the identification of a critical region with enhanced adaptability, and the swift recovery from intruder-induced disturbances all highlight the potential of minimal interaction models for swarm robotics. Our findings reinforce evidence from field experiments demonstrating that drones can maintain coherent motion in outdoor environments using only local sensing and interactions with a small subset of neighbors \cite{vasarhelyi2014, zhou2022science}. They also connect with emerging work on decentralized coverage, search-and-rescue operations, and GPS-denied navigation, where simple coordination primitives have proven effective under uncertainty \cite{horyna2023autonomous, saska2016gps}. The ability to tune a swarm’s position in the phase diagram provides a powerful mechanism to adjust its responsiveness and stability. This insight opens the door to adaptive controllers that combine minimal interaction rules with higher-level planning architectures, enabling drones to navigate complex, cluttered, or adversarial environments while preserving scalability. Overall, this work contributes a conceptual and practical foundation for designing robust and adaptable aerial swarms capable of collective intelligence in dynamic real-world settings.

\section*{Methods}
\label{Methods}

\subsection*{Drone platform and control architecture}

\begin{figure}
    \centering
    \includegraphics[width=\linewidth]{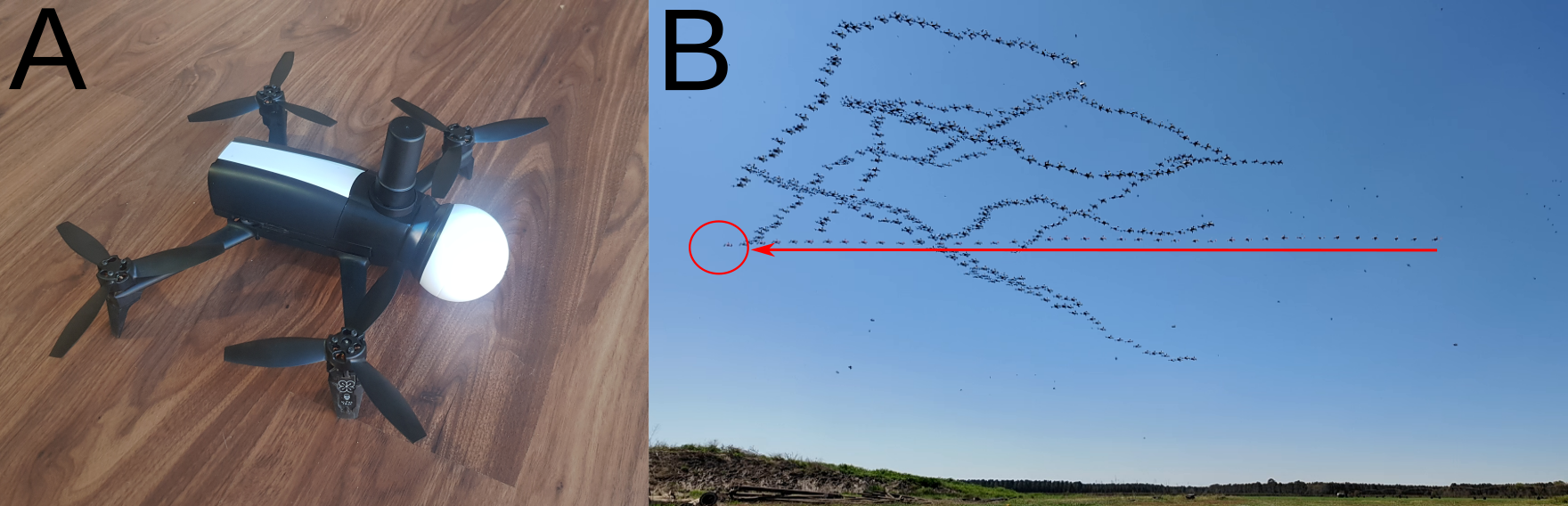}
    \caption{\textbf{Experimental drones}. (\textbf{A})~Parrot Bebop 2 drones modified by Dronisos with high-precision GPS modules and a front-mounted RGB light were used in all experiments. (\textbf{B})~Representative time sequence (10 seconds) of a flight involving a swarm of 10 drones and an intruder (highlighted by the red circle). The red arrow indicates the intruder’s trajectory as it crosses the swarm.}
    \label{fig:bebop2}
\end{figure}

The drone platform used in all experimental flights is a modified Parrot Bebop 2, custom-adapted by the drone show company Dronisos (Bordeaux, France). The vehicle is a 500-g quadrotor with a footprint of approximately 40$\times$40\,cm and a nominal flight endurance of about 25\,min (Figure~\ref{fig:bebop2}A). In this version, the original GPS module was replaced with a Ublox F9P receiver, enabling centimeter-level positioning through differential corrections provided by a ground reference station. This configuration yields a positional accuracy of roughly 10\,cm, which is essential for safe and coherent operation when drones fly in close proximity.

Swarm control is implemented within the ROS2 framework. Each drone executes its own controller in an independent node, but due to technical constraints associated with the proprietary Dronisos flight-control system, the vehicles are commanded remotely through the company’s ground-based control center. The control inputs consist of position-setpoint updates accompanied by an associated time-to-target value, ensuring that drones move at approximately constant velocity.

Simulations employ the same ROS2 architecture used for real flights, with the key difference that control commands are delivered to a custom flight-dynamics model rather than to physical drones. The simulator is implemented as an additional ROS2 node and reproduces the inertial and actuation constraints of the Bebop platform. For both real and simulated experiments, all state variables and control messages are logged using ROS2 Bag for subsequent post-processing and analysis.

\subsection*{Field experiments with a drone swarm and numerical simulations}

The outdoor experiments were conducted at the Dronisos testing facility located in Cestas, near Bordeaux, France. The proprietary Dronisos flight-control system was used during takeoff and landing and to enforce geofencing restrictions that prevent drones from leaving the authorized flight volume. Experimental flights took place on 9-10 April 2025 under wind conditions ranging from 18 to 22\,km/h with gusts reaching 33\,km/h.

Once airborne, the swarm controller was activated. It received position and velocity data from the Dronisos ground control station and returned position-setpoint commands that determined the drones’ next target locations. Telemetry was streamed at 1\,Hz, whereas control updates were issued at 2\,Hz. The swarm started moving immediately once the controller was launched and landing procedures were initiated through the Dronisos interface at the end of each flight. All flight data were recorded using the ROS2 logging system for subsequent analysis.

The swarm controller can be configured for multiple scenarios, numbers of drones, and intruders. Scenario-specific parameters—including alignment and attraction gains—can be updated automatically during flight. Each run was limited by battery capacity, typically 15–20\,min, and all scenarios were repeated at least twice. Some results aggregate the full dataset, whereas others highlight representative examples chosen for clarity.

Three experimental scenarios were performed. In all cases, the swarm flew within a virtual circular arena of 50\,m diameter (significantly larger than the mean free path of the drones) at altitudes between 5 and 15\,m.

(1) Baseline scenario without an intruder: 10 drones flew while the alignment gain was incremented automatically every 60\,s across a range spanning the swarming and schooling regimes with a $0.025$ increment on $\gamma_{\rm Ali}$. Group-level dispersion and polarization were recorded continuously.

(2) Switching-gain scenario: the same group of 10 drones alternated every 60\,s between the lowest and highest alignment gains, forcing repeated transitions between swarming and schooling. This enabled characterization of the dynamics and asymmetry of these phase transitions.

(3) Intruder scenario: the baseline experiment was repeated with an additional drone acting as an intruder. The intruder’s trajectory was autonomously directed toward the swarm barycenter (Figure~\ref{fig:bebop2}B). Once outside the arena boundary, the intruder paused briefly to allow the swarm to relax before performing a new approach.

All scenarios were reproduced in simulation using the same swarm controller and a flight-dynamics model calibrated from closed-loop identification of real flight trajectories. Intruder trajectories, consisting of sequences of straight segments, were used for model identification. A first-order response model captured the delay between a commanded and an achieved position:
\begin{equation}
    \tau\ \dot{x} + x = x_d
    \label{eqn:fish_model}
\end{equation}
where $x$ is the drone’s position, $x_d$ its desired position, and $\tau$ the time constant. 
Least-squares fitting (SciPy, Python) yielded 
$1.119$\,s. Simulations were run with the same parameters and durations as the corresponding outdoor flights.

\subsection*{Preprocessing of the drone trajectory data}

The downstream telemetry available at 1\,Hz is insufficient to compute the quantification parameters with the required accuracy. It is therefore necessary to preprocess the position data with three objectives in mind: (i) estimating position and velocity at a sufficiently high sampling frequency, (ii) synchronizing all trajectories from the drones and the intruder, and (iii) reducing measurement noise.

The raw dataset contains timestamped positions and orientations of the drones, together with their commanded positions. The first preprocessing step consists of applying an interpolation procedure that enables resampling the trajectories at the desired frequency while enforcing a common time vector across all time series. We use the \emph{Akima1DInterpolator} from the SciPy library, which offers a key advantage over linear interpolation by producing a $\mathcal{C} 1$ continuous function that passes through the original data points and can be analytically differentiated to obtain velocity estimates. The resulting velocity signals may still exhibit residual noise. When necessary, a Savitzky–Golay filter is applied to smooth the data without introducing temporal lag.

\subsection*{Computational Model}
\begin{figure}[ht]
    \centering
    \includegraphics[width=0.5\linewidth]{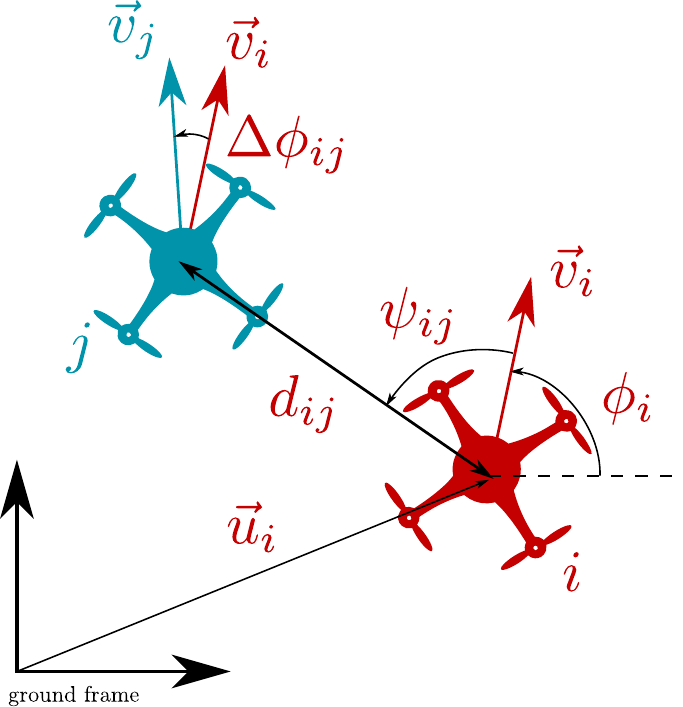}
    \caption{\textbf{State variables and relative position and orientations}. $\vec{u}_i$, $\vec{v}_i$ and $\phi_i$ are respectively the position, the velocity and the heading of drone $i$ relative the fixed ground frame. $d_{ij}$, $\psi_{ij}$ and $\Delta\phi_{ij}$ are respectively the distance, the viewing angle and the heading difference between drones $i$ and $j$. The heading corresponds here to the ground course given by the velocity vector. The yaw angle of the drone can be controlled independently and is not relevant in this case.}
    \label{fig:swarm_frame}
\end{figure}

When considering a swarm of $N$~UAVs flying in a 3D space, the position and velocity vectors of the UAV~$i$ are respectively $\vec{u}_i = (x_i, y_i, z_i)$ and $\vec{v}_i = (v_i^x, v_i^y, v_i^z)$. The instantaneous state of the UAV~$i$ is determined by the vector $\vec{U}_i(t) = (\vec{u}_i(t), \phi_i(t))$, where $\phi_i = {\rm atan2}(v_i^y,v_i^x)$ denotes the ground course angle (referred as heading) of the UAV in the horizontal plane. Then, the command vector can be written as $\vec{V}_i(t) = ({\rm v}_i(t), v_i^z(t), \omega_i(t))$, where ${\rm v}_i = \|(v_i^x,v_i^y)\|$ is the longitudinal speed and $\omega_i$ is the angular turning rate. The angles and relative positions and orientations are defined in Figure~\ref{fig:swarm_frame}.

Without border conditions, the agent's speed and heading variation $\delta \vec{V}_i$ is updated at a fixed frequency according to the social interactions with other agents (${\rm soc}$), the operational goal that includes an attraction to a given altitude (${\rm nav}$), and eventually a given direction ($\delta\phi_i^{\rm nav}$):
\begin{align}
\delta {\rm v}_i & = \delta {\rm v}_i^{\rm soc} + \delta {\rm v}_i^{\rm nav},
\label{eqn:speed_cohesion_fish}
\\
\delta v_i^z & = (\delta v_i^z)^{\rm soc} + (\delta v_i^z)_\parallel + (\delta v_i^z)^{\rm nav} + (\delta v_i^z)^{\rm intruder},
\label{eqn:speed_vertical_fish}
\\
\delta\phi_i & = \delta\phi_i^{\rm soc} + \delta\phi_i^{\rm nav} + \delta\phi_i^{\rm intruder}.
\label{eqn:fish_more}
\end{align}
The social terms are $\delta {\rm v}_i^{\rm soc} = \sum_{j\in J} \delta {\rm v}_{ij}$, $(\delta v_i^z)_{\rm soc} = \sum_{j\in J} \delta v_{ij}^z$, and $\delta\phi_i^{\rm soc} = \sum_{j\in J}(\phi_{ij}^{\rm Ali} + \phi_{ij}^{\rm Att})$, where $J$ is the set of agents with which agent~$i$ interacts (typically~1 or~2). They are given by pairwise functions that describe the effect of the longitudinal speed, the vertical speed, and the combination of alignment and attraction, respectively:
\begin{align}
\label{eqn:speed_one_uav_inf}
    \delta {\rm v}_{ij} = &\ \gamma_{\rm Acc} \cos(\psi_{ij})
    \big(d_0^{\rm v}-d_{ij}^{\rm c}\big)
    \left(1+\frac{d_{ij}^{\rm c}}{l_{\rm Acc}}\right)^{-1},
\\
\label{eqn:vert_one_uav}
    \delta v_{ij}^z = &\ \gamma_{\rm z} \tanh \left(\frac{d_{ij}^{\rm z}-d_0^z}{a_z}\right) \exp \left[-\left(\frac{d_{ij}^{\rm c}}{l_{\rm z}} \right)^2\right],
\\
\label{eqn:ali_one_uav}
    \delta\phi_{ij}^{\rm Ali} = &\ \gamma_{\rm Ali}  \sin(\Delta\phi_{ij}) \big(1+\alpha_{\rm Ali}\cos(\psi_{ij})\big)
    (d_{ij}^{\rm c}+d_0^{\rm Ali})
    \exp{\left[-\left(\frac{d_{ij}^{\rm c}}{l_{\rm Ali}} \right)^2\right]},
\\
\label{eqn:att_one_uav}
    \delta\phi_{ij}^{\rm Att} = &\ \gamma_{\rm Att} \sin(\psi_{ij}) \big(1 - \alpha_{\rm Att} \cos(\psi_{ij})\big) 
    (d_{ij}^{\rm c}-d_0^{\rm Att}) \left[1+\left(\frac{d_{ij}^{\rm c}}{l_{\rm Att}}\right)^2 \right]^{-1}.
\end{align}

These social interaction functions depend on four variables characterizing the geometric relative state of pairs of agents~$ij$: {\bf [i]} the angle $\psi_{ij}$ with which~$i$ perceives~$j$, {\bf [ii]} their heading difference $\Delta\phi_{ij} = \phi_j - \phi_i$, {\bf [iii]} the vertical separation $d_{ij}^{\rm z}$, and {\bf [iv]} the distance $d_{ij}^{\rm c}$ between them. The angles $\psi_{ij}$ and $\Delta\phi_{ij}$ are calculated with respect to the projected position of agent~$j$ onto the local horizontal plane of agent~$i$. The distance between agents $d_{ij}^{\rm c}$ is weighted by amplifying the vertical axis with a coefficient $\sigma_z$ in order to prevent collisions or failures between vertically aligned agents due to the columns of air perturbations generated by multirotors:
\begin{align}
\label{eqn:d_coupled}
    d_{ij}^{\rm c} = \sqrt{(x_i-x_j)^2 + (y_i-y_j)^2 + \left(\frac{z_i-z_j}{\sigma_z}\right)^2}.
\end{align}
Additionally,  alignment and attraction terms can integrate a wall interaction, function of the horizontal distance to the border of the flying space. Its impact is to create a repulsion force  and to reduce the intensity of social interactions when the agent is closer to the border, as presented in~\cite{verdoucq_bio-inspired_2022}. For clarity, this effect is not included in equations \ref{eqn:ali_one_uav} and \ref{eqn:att_one_uav}.

The parameters characterizing the scale and range of action of these forces are the distances from equilibrium $d_0^{\rm v}$, $d_0^z$, $d_0^{\rm Ali}$, $d_0^{\rm Att}$, and their respective ranges of action, determined respectively by $l_{\rm Acc}$, $l_{\rm z}$, $l_{\rm Ali}$, and $l_{\rm Att}$. Note that the attraction term (eq. \ref{eqn:att_one_uav}) changes sign at the distance $d_0^{\rm Att}$, which means that it becomes repulsive when the distance to neighbor is smaller than that. It is the only collision avoidance mechanism that we are considering in this study. If more sophisticated solutions can be found for operational application, our goal is too not disturb the natural behavior of the swarm suggest to our social interactions model.

Recent findings in social fish species have shown that individuals acquire a minimal amount of information about a limited number of neighbors (one or two) to coordinate movements at the group level~\cite{jiang2017identifying,lei2020computational}, thus facilitating decision-making processes and preventing cognitive overload. This feature is included in our model by selecting the number and identity of the agents to interact with (two agents in our study), reducing the amount of information that each UAV must process~\cite{verdoucq_bio-inspired_2023}. The interacting neighbors are precisely those that exert the highest influence on the focal agent, where the influence that agent~$j$ exerts on agent~$i$ is given by

\begin{align}
\label{eqn:total_int}
I_{ij}(t) = \sqrt{(\delta {\rm v}_i)^2 + (\delta v_i^z)^2 + (\delta \phi_i {\rm v}_i)^2}.
\end{align}

The operational terms $\delta {\rm v}_i^{\rm nav}$, $(\delta v_i^z)^{\rm nav}$, and $\delta\phi_i^{\rm nav}$ serve to determine the guiding strategy for the swarm to adopt a given position, speed, and heading. Here, only $(\delta v_i^z)^{\rm nav}$ and $\delta\phi_i^{\rm nav}$ are used to attract drones to a specified altitude~$z_{\rm alt}$ and to a desired direction:
\begin{align}
\label{eqn:vertical_nav}
    (\delta v_i^z)^{\rm nav} =& (\delta v_i^z)_\perp = \gamma_{\perp} \tanh \left(\frac{z - z_{\rm alt}}{a_{\rm z}}\right) ,\\
\label{eqn:yaw_nav} 
    \delta\phi_i^{\rm nav} =& \gamma_{\rm nav} \sin(\Delta\phi_i^{\rm nav}),
\end{align}
where $\gamma_{\perp}$ and $\gamma_{\rm nav}$ control the intensity and $a_{\rm z}$ is the vertical attraction distance. These navigation terms can be neglected for free-flight scenarios. During experiments and simulations, the target altitude $z_{\rm alt}$ was set to 10\,m and the term $\delta\phi_i^{\rm nav}$ was not used.

Regarding terms related to intruder avoidance, we implemented a yaw interaction inspired by the shape of the repulsive force exerted by the wall in the horizontal plane, with adapted parameters. For vertical avoidance, the function is the repulsive part of the $\delta v_{ij}^z$-interaction. So, when close to an intruder~$k$, an agent~$i$ computes its avoidance behavior as follows:
\begin{align}
\label{eqn:yaw_rep}
    \delta \phi_{ik} = & \gamma_{\rm Rep} \sin(\psi_{ik}) \big(1+\cos(\psi_{ik})\big) \exp  \left[-\left(\frac{d_{ik}}{l_{\rm Rep}} \right)^2\right],
\\
\label{eqn:vert_rep}
\delta v_{ik}^z = & \gamma_{\rm z} \left[\tanh \left(\frac{d_{ik}^{\rm z}}{a_{\rm z}}\right)\right] \, \exp \left[-\left(\frac{d_{ik}^{\rm c}}{l_{\rm z}} \right)^2\right],
\end{align}
where $\gamma_{\rm Rep}$ is the horizontal repulsive intensity and $l_{\rm Rep}$ is the horizontal interaction distance with the intruder.
Finally, $(\delta v_i^z)_\parallel = \gamma_\parallel \sin({v_i^z/{\rm v}_i})$ is used to smooth abrupt changes in vertical position, also using a lower bound for ${\rm v}_i$.

The expressions (\ref{eqn:speed_one_uav_inf})-(\ref{eqn:total_int}) are the generalization to a 3D framework of the functions introduced in the 2D model~\cite{verdoucq_bio-inspired_2022}.
The values of the model's parameters used in the simulations and  in actual drone flights are given in Table~\ref{tab:parameters}.

\subsection*{Quantification of collective dynamics and responsiveness}

In this study, we use three standard metrics from collective-motion research to quantify the structure of the swarm and its response to perturbations:

\begin{itemize}

    \item \textbf{Group dispersion}, which measures how tightly individuals cluster around the swarm barycenter~$B$:
    \begin{align}
        \label{eqn:dispersion}
        D(t) = \sqrt{\frac{1}{N} \sum_{i=1}^{N} \| \vec{u}_i - \vec{u}_B \|^2 }.
    \end{align}
    A low dispersion indicates a compact, cohesive group, whereas larger values reflect spatial expansion or partial fragmentation.

    \item \textbf{Polarization}, which captures the degree of alignment among drones:
    \begin{align}
        \label{eqn:polarization}
        P(t) = \frac{1}{N} \left\| \sum_{i=1}^{N} \vec{e}_i(t) \right\| \in [0,1],
    \end{align}
    where $\vec{e}_i = \vec{v}_i / \lVert \vec{v}_i \rVert$ is the unit vector of drone~$i$'s instantaneous velocity. A value of $P(t)=1$ corresponds to perfect alignment, disordered motion (randomly aligned agents) yields values around $P_0 = (1/2)/\sqrt{\pi/N}$ ($P_0 \approx 0.28$ for $N=10$, $P_0 \approx 0.2$ for $N=20$). Values smaller than $P_0$ denote specific configurations where agents' orientations cancel each other.

    \item \textbf{Polarization susceptibility}, which quantifies fluctuations in group alignment under fixed control parameters:
    \begin{align}
        \label{eqn:pol_fluctuation}
        \chi_P = N \left[ \langle P^2 \rangle - \langle P \rangle^2 \right].
    \end{align}
    Higher values of $\chi_P$ signal greater sensitivity to perturbations and are typically associated with transitional or near-critical collective regimes.

\end{itemize}

\section*{Data Availability}

The datasets generated and/or analyzed during the current study are available in the Zenodo repository, https://doi.org/10.5281/zenodo.17902132.

\section*{Acknowledgments}
This work was supported by a grant from the University of Toulouse and the Région Occitanie/Pyrénées-Méditerranée to M.V., G.H. and G.T.. G.T., R.E., and C.S.~were supported by the Agence Nationale de la Recherche (ANR-20-CE45-0006-1). D.T.~has received funding from the European Union’s Horizon 2020 research and innovation program under the Marie Skłodowska-Curie grant agreement No.101154645. Special thanks go to Dronisos, Bordeaux, for their active support and availability for outdoor experiments.

\section*{Author Contributions}

M.V., G.T. and G.H. designed the research. M.V., G.T., C.S., R.E. and G.H.~developed the model. M.V.~implemented the software architecture for experiments and simulations. M.V., D.T. and G.H.~performed the experiments. D.T., G.T. and G.H.~analyzed and interpreted the experimental data. M.V., D.T., G.T. and G.H.~wrote the article. All authors read and approved the final manuscript

\section*{Competing Interests}

The authors declare no competing financial or non-financial interests.

\bibliography{references}
\bibliographystyle{naturemag}

\begin{table}[ht]
    \centering
    \includegraphics{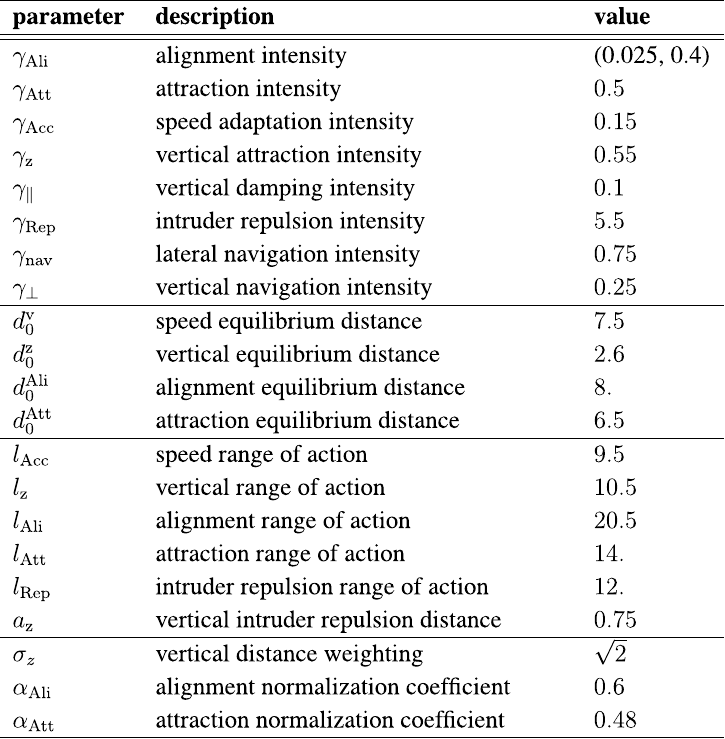}
    \caption{\textbf{List of model parameters}. The same parameter values were used in both the simulations and the field experiments.}
    \label{tab:parameters}
\end{table}

\end{document}